%% file: main.tex
\documentclass[conference]{IEEEtran}
\usepackage{cite}
\usepackage{graphicx}
\usepackage{textcomp}
\usepackage{xcolor}
\usepackage{url,afterpage,amssymb,amsfonts}
\usepackage{enumitem,makecell}
\usepackage{booktabs}
\usepackage{multirow}
\usepackage{subfig}
\usepackage{graphicx}
\usepackage{amsmath}
\usepackage{verbatim}
\usepackage{dblfloatfix}
\usepackage{color}
\usepackage{colortbl}
\graphicspath{{./img/}}
\usepackage{listings}
\usepackage{soul}
\usepackage{enumitem,makecell}
\usepackage{colortbl}
\usepackage{float}
\usepackage{wrapfig}
\usepackage{pgfgantt}
\usepackage[thinc]{esdiff}
\usepackage{algorithm,algpseudocode}
\algrenewcommand\textproc{}
\usepackage{tabularx,booktabs}
\usepackage{xcolor}
\newcolumntype{C}{>{\centering\arraybackslash}X} 
\makeatletter
\newenvironment{breakablealgorithm}
  {
   \begin{center}
     \refstepcounter{algorithm}
     \hrule height.8pt depth0pt \kern2pt
     \renewcommand{\caption}[2][\relax]{
       {\raggedright\textbf{\fname@algorithm~\thealgorithm} ##2\par}%
       \ifx\relax##1\relax 
         \addcontentsline{loa}{algorithm}{\protect\numberline{\thealgorithm}##2}%
       \else 
         \addcontentsline{loa}{algorithm}{\protect\numberline{\thealgorithm}##1}%
       \fi
       \kern2pt\hrule\kern2pt
     }
  }{
     \kern2pt\hrule\relax
   \end{center}
  }
\makeatother

\newcounter{savealgorithm}
\newenvironment{subalgorithms}
 {%
  \stepcounter{algorithm}%
  \edef\currentthealgorithm{\thealgorithm}%
  \setcounter{savealgorithm}{\value{algorithm}}%
  \setcounter{algorithm}{0}%
  \renewcommand{\thealgorithm}{\currentthealgorithm\alph{algorithm}}%
 }
 {%
  \setcounter{algorithm}{\value{savealgorithm}}%
 }

\def\BibTeX{{\rm B\kern-.05em{\sc i\kern-.025em b}\kern-.08em
    T\kern-.1667em\lower.7ex\hbox{E}\kern-.125emX}}

\begin{document}

\title{PFSL: Personalized \& Fair Split Learning with Data \& Label Privacy  for thin clients}

\author{\IEEEauthorblockN{Manas Wadhwa, Gagan Raj Gupta, Ashutosh Sahu, Rahul Saini, Vidhi Mittal}
\IEEEauthorblockA{\textit{Department of Electrical Engineering and Computer Science} \\
\textit{Indian Institute of Technology, Bhilai}\\
 Bhilai, India \\
\{manasw, gagan, ashutoshsahu, rahuls, vidhimittal\}@iitbhilai.ac.in}
}

\maketitle
\begin{abstract}
The traditional framework of federated learning (FL) requires each client to re-train their models in every iteration, making it infeasible for resource-constrained mobile devices to train deep-learning (DL) models. Split learning (SL) provides an alternative by using a centralized server to offload the computation of activations and gradients for a subset of the model but suffers from problems of slow convergence and lower accuracy. 

In this paper, we implement PFSL, a new framework of distributed split learning where a large number of thin clients perform transfer learning in parallel, starting with a pre-trained DL model without sharing their data or labels with a central server. We implement a lightweight step of personalization of client models to provide high performance for their respective data distributions. Furthermore, we evaluate performance fairness amongst clients under a work fairness constraint for various scenarios of non-i.i.d. data distributions and unequal sample sizes. Our accuracy far exceeds that of current SL algorithms and is very close to that of centralized learning on several real-life benchmarks. It has a very low computation cost compared to FL variants and promises to deliver the full benefits of DL to extremely thin, resource-constrained clients. 

\end{abstract}

\begin{IEEEkeywords}
Distributed Machine Learning, U Shaped Split Learning, Federated Learning, non-i.i.d., Personalization, Fairness, Image Classification
\end{IEEEkeywords}

\input{sections/Introduction}
\input{sections/RelatedWork}
\input{sections/SystemModel}
\input{sections/algorithms.tex}

\input{sections/Experiments_and_results.tex}
\input{sections/Conclusion}

\bibliographystyle{IEEEtran}
\bibliography{ref}
\newpage
\input{sections/Appendix.tex}

\end{document}

%% file: sections/Introduction.tex
\section{Introduction}\label{sec:intro}
Data is created at the edge and often owned by individuals or groups who take data privacy seriously. The exponentially growing data usage in deep learning (DL) \cite{big-data} has triggered massive collections of user (often anonymous) data by large organizations leading to concerns about data privacy \cite{split_med,ml_priv,ml_priv2}. Thus, it is desirable to develop high-performance distributed machine learning (DML) systems that run on low-cost,  thin clients without compromising data privacy. 

In Federated Learning (FL) \cite{split_comp_fed, split_fed, fed_learn_rev, fed_learn_med} each client receives a base model from a secure central server, and trains and updates the model parameters according to its own personal data. These updated models generated by a number of clients are then aggregated securely \cite{fed_learn_rev, fed_priv} resulting in a more accurate and generalized model. The FL architecture makes it particularly difficult to deploy DL models on resource-constrained (thin) clients such as mobile phones, sensors, smart cameras, etc. due to their high memory, storage, computing, and energy requirements during training \cite{HW_Eval}. 

The split learning (SL) framework \cite{split_main}, provides a promising alternative for thin clients. It splits the DL model into parts and offloads resource-intensive parts to a central server. The clients are responsible for updating the parameters only in the parts of the model that are owned by them, while the server is responsible for updating parameters for its part of the model. \cite{HW_Eval} has raised concerns such as slow training rate, low accuracy, and high resource utilization with SL for thin clients. Most of the papers in split learning \cite{split_comp_fed, split_fed, split_fed_edge_iot, split_guard, split_label_leak, split_label_protect, split_locfedmix, split_med, split_no_peek, split_scale} have only focused on a two-stage architecture where input data remains with the clients and the labels are shared with the server. For many applications, labels may reveal sensitive information such as diseases present, travel history, purchase history, search history, etc. for each client and thus can't be shared. 

\begin{figure}[h]
\includegraphics[width=\linewidth]{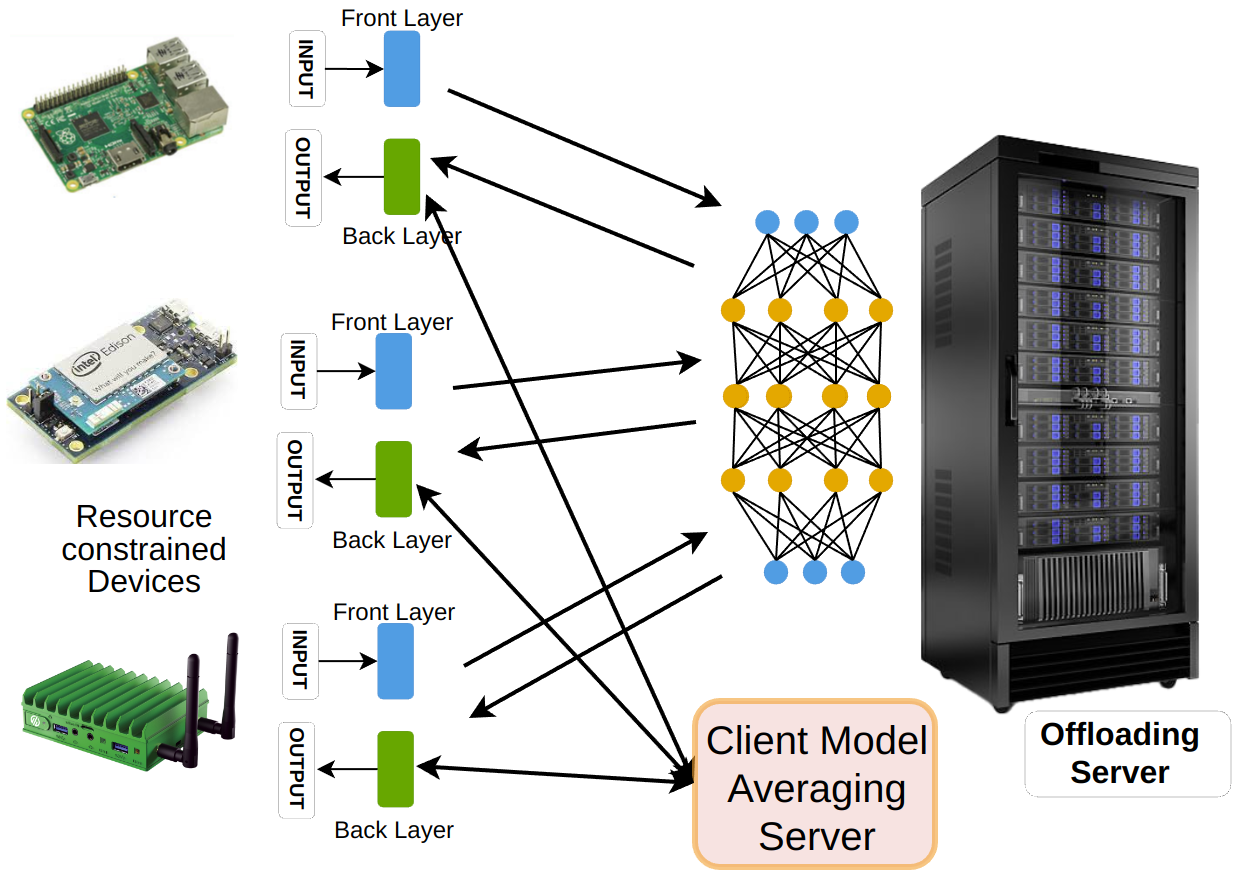}
\caption{PFSL framework supports training DL models on resource-constrained devices by off-loading heavy computation to an offloading server. Data and its labels reside with the clients and the clients only train a small portion of the DL model.
  \vspace{-0.5em}} 
\label{fig: fig2}
\end{figure}

In this paper, we develop a three-stage architecture (Fig \ref{fig: fig2}) where the server is simply an offloading device with no information whatsoever about the applications that the clients are running or the labels with the clients. Moreover, for aggregation of the client-side layers, we employ a separate secure client model averaging server. Thus as opposed to traditional two-stage split learning, the servers here do not have access to labels, thus making the model inversion attacks \cite{Model-Inversion, MI2} etc difficult.  

We make the following contributions through our Personalized Fair Split Learning (PFSL) framework which enables even \textbf{thin clients} to realize their goal of training \textbf{personalized DL models}: 

\begin{itemize}
\item{\textbf{Enable Parallel U-Shaped SL with Label privacy} by implementing a highly scalable, three-stage SL architecture, which doesn't require the clients to share labels of their data with the servers (Fig \ref{fig: fig2}). This was proposed in \cite{split_main} but ours is the first distributed/parallel implementation and detailed evaluation on real-life benchmarks. 
}
\item{\textbf{Adapt Transfer Learning } to minimize the resources needed on the client's end to achieve an accurate model. With detailed experiments on real-life benchmarks, we demonstrate higher accuracy than standard FL \cite{fed_learn}, FL with transfer learning \cite{FL-TL} and SL variants \cite{split_fed} and faster convergence minimizing computation and communication overheads.}

\item{\textbf{Personalized SL} for meeting the specific objectives such as personalization, and generalization of clients models. For example, one client (pathologist) may want their model to be more accurate for the images captured by the medical device in their lab (personalization). Another client (large pathology lab) may want their model to work well across all of the devices (generalization).}

\item{\textbf{Fairness:} Implemented and evaluated ``performance fairness'' under a work fairness constraint which ensures that the clients perform a similar amount of work, regardless of the amount of data they possess. In contrast, most FL and SL variants require clients with more data to work more during training. Furthermore, when compared to existing approaches, the performance fairness (measured as the standard deviation of test accuracy across clients) under PFSL is consistently lower.}

\item{\textbf{Reproducible Code:} To facilitate research in SL by the community, we have developed a modular, high-performance GPU-accelerated implementation of our framework\cite{Manas_Wadhwa_and_Gagan_Gupta_and_Ashutosh_Sahu_and_Rahul_Saini_and_Vidhi_Mittal_PFSL_2023}.  This allows us full-scale distributed learning system simulation, in which hundreds of thin clients can train their models. Our framework allows quick convergence based on global validation accuracy. Furthermore, a client can enter and leave the system whenever they wish to do so. This is realistic for thin, edge devices deployed remotely where the network connections may not persist for too long. 

}
\end{itemize}

In our framework, any DL model is split into three parts: Client-Front, Client-Back, and Server-Central layers as shown in Fig \ref{fig: fig2}. The number of layers in any of the parts is configurable, depending on the complexity of the task and the resources available to the clients. The number of client layers can be kept small (1 or 2) to ensure that highly resource-constrained thin clients can train their models using our framework. The weights of some of the layers can be frozen if they are obtained from a pre-trained model and the weights of unfrozen layers can be updated (transfer learning). 

In \cite{Jiang19-meta}, the authors state three objectives for FL: solid initial model, improved personalization models,  and fast convergence. We have designed PFSL to achieve all three by starting from a pre-trained model and focusing on its training for achieving high performance on general test distribution. This is critical to achieving a solid initial model and fast convergence. Furthermore, we have adapted standard approaches of personalization in FL \cite{P1, P2, P3, P4} to SL by implementing a lightweight step for the personalization of client layers which is suitable for thin clients. Instead of implementing personalization in a multi-task framework as in \cite{ditto_main} which is very resource intensive, we make it coarse-grained by first running a \emph{``generalization''} phase for all clients and introducing an additional \emph{``personalization''} phase to boost their accuracy on their personal data distribution. Here, we use early stopping \cite{EarlyStopping} for regularization. We demonstrate through detailed experiments, that both the clients with more data points and fewer data points can achieve significant performance gains while working equitably.

We present relevant background and discuss related work in Sec \ref{sec:rel_work}. A complete description of our PFSL Framework and relevant notations are presented in Sec \ref{sec:background_sys_model}. We describe our algorithms in detail in Sec \ref{sec:Algorithms}, evaluate their performance and compare with the state-of-the-art in Sec \ref{sec:experiments_and_results}, and draw conclusions in Sec \ref{sec:conclusion}.

%% file: sections/RelatedWork.tex
\section {Background and Related Works}
\label{sec:rel_work}

Centralized Training requires all the data from different sources to be pooled together at a central server on which training is performed raising concerns about data privacy. There are several lines of work on various attacks on ML Models\cite{ref_fed_privacy, ref_fed_diff_private, ref_break_privacy_fl, Model-Inversion, MI2} even when they have been trained on anonymized data. Distributed Machine Learning (DML) methods like Federated Learning (FL) \cite{fed_incep} and its variant, Split Learning (SL) \cite{split_main} were introduced to overcome these limitations of centralized learning. FL and SL attempt to limit sharing of raw data and offer to train a robust and generalized model through collaborative learning techniques. From the vast and ever-growing literature on FL and SL, we have selected a few topics to provide the reader with enough background to fully understand the working of PFSL and our contributions. For ease of comparison between different distributed machine learning algorithms refer to Table \ref{tab: layer_dis}. 

\begin{table*}
    \centering
    \begin{tabular}{|l|l|l|l|l|}
    \hline
        \textbf{Algorithm } & \textbf{Front Layers } & \textbf{Central Layers } & \textbf{Back Layers } & \textbf{Parallel / Sequential} \\ \hline
        FL \cite{FL-TL} & Client & Client  & Client  & Parallel\\ \hline
        SL \cite{split_main} & Client   & Server   & Server  & Sequential \\ \hline
        SFLv1 \cite{split_fed} & Client   & Server   & Server  & Parallel  \\ \hline
        SFLv2 \cite{split_fed} & Client   & Server   & Server   & Sequential \\ \hline
        PFSL (this paper)& Client    & Server   & Client   & Parallel  \\ \hline
    \end{tabular}
    \caption{A summary of various Federated and Split Learning approaches with respect to the distribution of neural network's layers between a client and server and whether multiple clients are trained sequentially or in parallel.}
    \label{tab: layer_dis}
\end{table*}

\subsection{\textbf{Federated Learning}}
In FL, each client initially loads a central model from the server. Once loaded, the model will train and update the model parameters according to its local private data. Thus a different variant of the local model is generated. Once this part is done, the weights and other model-specific information are uploaded to the central server. These updated models generated by a number of clients are then aggregated securely \cite{fed_learn_rev, fed_priv} resulting in a more accurate and generalized model. Through multiple epochs/rounds of local training and secure aggregation, the accuracy of the model gradually improves and converges. 

As shown in Table \ref{tab: layer_dis}, federated learning requires the client to handle all the workload during training.  Considering the size and complexity of modern DNNs, federated learning requires the client to do lots of computations, indicating that it is unsuitable for resource-constrained devices. In Split learning, only the front layers are with the clients. Thus a client in the SL framework has to take care of the forward pass, backward pass, and weights updates of only the front layer.

Over the years, there has been tremendous interest and progress in optimizing FL for various objectives such as fairness, reduction in communication costs, non-i.i.d. data distribution across clients, mitigating attacks from malicious clients, etc.  \cite{fed_non_iid, split_fed, fed_learn_rev, fed_learn_med, split_label_leak}. FL requires the devices to train their models locally, which is very challenging on limited-resource hardware devices \cite{fed_learn, fed_learn_rev, fed_priv, HW_Eval}. As the State-of-the-Art Deep Learning models get more complex and sophisticated, it becomes highly infeasible for implementation on low-end Internet of Things (IoT) devices \cite{HW_Eval}. 

\subsection{\textbf{Split Learning}}
Some of these limitations were addressed by split-learning \cite{split_main, split_med} which proposed the different ways of splitting DL models, data-partitioning schemes, and its novel use cases. However, its implementation was sequential, required heavy synchronization during training, and the overheads were not quantified clearly. Subsequent research focused on scalability and faster convergence \cite{split_locfedmix, split_scale}, amalgamation with FL \cite{split_comp_fed, split_fed, split_fed_edge_iot}, increasing data privacy \cite{split_guard, split_label_leak, split_label_protect, split_locfedmix, split_med, split_no_peek} but only for the two-stage architecture where labels are shared with the server. There are two important variants of two-stage architecture: Sequential Split Learning (SSL) and Parallel Split Learning (PSL). 

In SSL, the clients are presented with the front part of the model, and the remaining model along with all data labels in the training set are shared with a central trusted server. The clients are selected sequentially for training and pass on their activations to the server. The server completes the forward pass, computes loss, and returns gradients to the clients via back-propagation. This process continues sequentially for all clients and the gradients/weights on the server side are merged. SSL achieves high accuracy \cite{split_scale} as the number of clients increases providing the desired benefits of DML. In PSL, the client and server operations are similar to SSL, with the distinct difference that multiple clients are connected to the server simultaneously. This method helps achieve good results with lower latency on the client side and utilizes parallel computing for training and testing the model \cite{split_scale}.

The three-part architecture (also called U-shaped Learning) provides both data and label privacy but is not well explored. We found that the implementation of U-shaped Learning is much more complex as it requires the clients to train additional (last few) layers, compute loss and pass gradients back to the server, thus increasing their computation and communication overhead. By utilizing publically available pre-trained models \cite{aws-tl} in PFSL, we have reduced the computation and communication load on the clients considerably (detailed in Sec 
\ref{sec:experiments_and_results}).  


\subsection{\textbf{Personalization}}
The heterogeneity of data at the clients presents both a challenge and opportunity for DML. It is a challenge as the diversity makes it difficult to get convergence to optimal weights/parameters for the model. It is an opportunity if the clients are allowed to develop personalized models that are optimized for their particular data distributions. In \cite{Smith17}, personalized FL was approached via a primal-dual multi-task learning framework for convex Loss functions. Subsequent works have applied clustering of clients with similar data distributions \cite{Ghosh20, Sattler20}, fine-tuning or transfer learning for personalization \cite{Zhao18, Yu20}. 

\subsection {\textbf{Fairness}}
When the clients' data distributions are non-i.i.d. and their sample sizes are heterogenous it is likely that there will be variance in their test accuracy \cite{HashimotoSNL18}. Thus, the goal of ``Performance Fairness'' is to \textbf{minimize std. deviation} of test accuracy \cite{li2019fair} while maintaining good average test performance. \cite{Mohri19} have used  minimax optimization to encourage uniform test performance across clients. In \cite{ditto_main}, a strong case was made that fine-grained personalization inherently provides the benefits of fairness and robustness to data and model poisoning attacks. However, it is very resource intensive for thin clients as they have to maintain two models and perform forward pass and back-propagation on both, doubling the costs.\cite{ditto_main, P1, P2, P3, P4} also suggest a lightweight coarse-grained personalization scheme which we have adapted in this paper.

As explained before, we add a \textbf{work equality constraint} in each round in PFSL by ensuring that all clients do an equal amount of work during training, regardless of the number of data samples possessed by them. If a client may participate in more rounds to get more accurate models and we leave that decision to the clients.

%% file: sections/SystemModel.tex
\section{System Model}
  \vspace{-0.5em}
\label{sec:background_sys_model}
\begin{figure}[h]
\includegraphics[width=\linewidth]{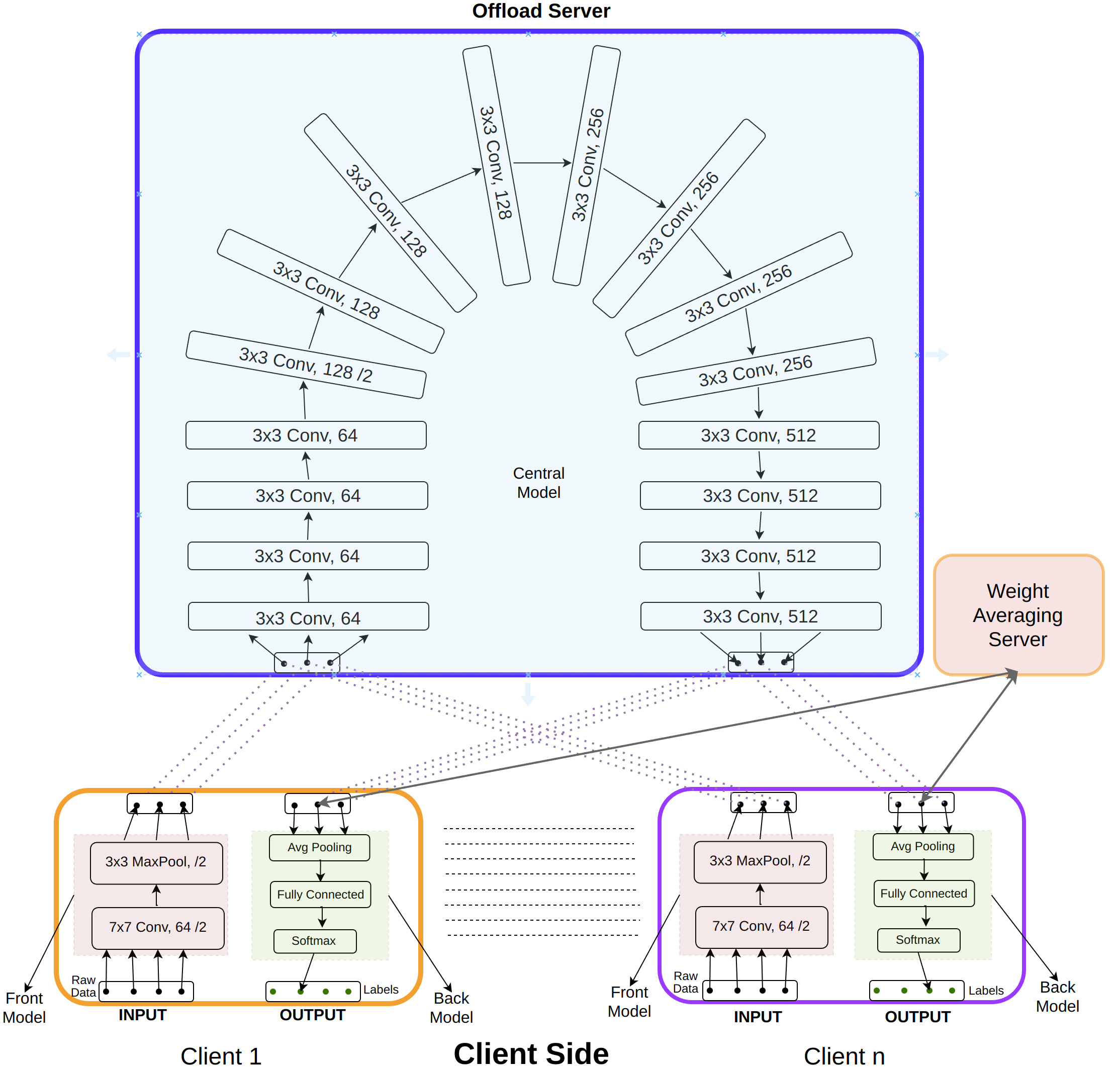}
\caption{PFSL System Architecture using ResNet18 as an example for splitting the model into the front, central, and back models for training.} 
\label{fig:mdl_main}
\end{figure}
Deep Learning (DL) models, by definition, are neural networks with 10s to 100s of layers, with each layer comprising 10s to 100s of neurons. An example is ResNet-152 \cite{ResNet-152}, which has 152 layers, 60M parameters, and achieved state-of-the-art accuracy on the ImageNet \cite{ImageNet} benchmark. Training these models requires a tremendous amount of computation and energy \cite{HW_Eval} which is not feasible for thin clients. Our first architecture choice in PFSL, therefore, is to split the model into three parts: front, middle, and back. We illustrate in Fig \ref{fig:mdl_main}, with an example of ResNet-18, how a model can be split. In this figure, the first 2 and last 2 layers reside with the client and the remaining 14 with the offloading server. The middle layers are usually much heavier and we make the offloading server responsible for the bulk of the computations. Furthermore, by starting with pre-trained DL models, we freeze the weights of the front layers at the clients, thereby reducing the burden of training even further. The \emph{offloading server} only receives activations and gradients from the clients and thus has no knowledge of the client applications. Also, as shown in the figure, we keep the \emph{client model averaging server} separated from the offloading server, thus preventing any information leak of the layers or the model architecture with the clients. Therefore, this method is useful for preserving privacy by preventing model inversion attacks and easing the computational burden on thin clients.

The entire training process is divided into multiple rounds.  During a round, a certain number of clients participate in the training process and stay for the entire duration. We also handle cases where some drop out due to disconnects, low battery, etc. by incorporating time-outs which ensure that the model continues to be updated with the remaining clients. Within a round, all the clients participate in the generalization phase (1) of our framework consisting of multiple global epochs.  Within a global epoch, all clients perform their local epochs in parallel. A global epoch will be considered completed when all the clients finish their local epoch. In a local epoch, we iterate over the local data of a client by forming batches. After the completion of a round, the clients may choose to continue to the personalization phase (2) of our framework or may choose to leave the system. 

We note that PFSL allows multiple clients ($n$, configurable) to train in parallel. To maintain correctness, it is essential to create $n$ copies of the Central model to store activations for each client so that the gradients can be calculated in parallel during the back-propagation phase. To support a large number of clients, the server capacity can be scaled by running server instances (micro-services) in the cloud. We provide full details of our remote procedures (APIs) in Sec \ref{sec:Algorithms}, but we don't discuss the cloud implementation further in this paper and focus only on the DML aspects.

\subsection{Work Fairness}

We introduce a novel \textbf{work fairness} constraint in this paper which ensures that all the clients who participate in a global epoch while training, perform an equal amount of work. It would be unfair for a client with more training data to do more work than the clients with less training data, which also achieve similar performance. Since we are dealing with \textbf{thin clients}, it is important to understand the work done by a client in the entire training process. More formally, we define below the \textbf{work done by the $ith$ client} in the training process. 

\begin{equation}
\label{eqn:work_fairness}
        O_i = E_{gc} * [(d_i/b) * C_i]   
  \vspace{-0.5em} 
\end{equation}

Here, $(d_i/b)$ is the number of iterations per client in one global epoch, $E_{gc}$ is the number of global epochs of the system for convergence, and $C_i$ is the compute operations of $ith$ client per iteration.

All clients in our framework perform the same number of iterations per global epoch regardless of the data with them i.e, $(d_i)/b = \min_{\forall i \in Clients} (d_i)/b$. The data of any client that is not used for training in a particular global epoch is carried over to the next global epoch. This is not imposed in other algorithms, which makes the client with more data do more work in training.


In our framework, a client has lightweight front and back layers. As we do not update the weights of the front layers, a client only has to do a forward pass through it, and the back layers do forward and backward passes and weight updates. To calculate the work done by $ith$ client during the entire training, we also multiply $C_i$ by the number of iterations and the number of epochs till convergence. Since our algorithm achieves convergence very quickly (refer to experiments), $E_{gc}$ will be very less; thus, the work done by the $ith$ client in our algorithm is lesser than other algorithms for the same setting (Table \ref{tab : wfair}). 


    





%% file: sections/algorithms.tex
\section{Algorithms}\label{sec:Algorithms}
In this section, we describe our algorithms for the training of the models for the personalization goal of clients. We will highlight the steps which have been designed to ensure high performance. As mentioned before, in PFSL, there are three types of entities: clients, offloading server, and client model averaging server. For ease of understanding, we have presented the relevant details of the algorithms run by each of them. These can also be thought of as APIs (and their implementation details) supported by these entities to coordinate with each other. Algorithm 1a describes the API provided by the secure client model averaging server that is responsible for averaging the weights of the client models calling it remotely. Each client participating in the training will run the APIs provided in Algorithm 1b. Offloading Server instances will run the APIs in Algorithm 1c.

\begin{table}[H]
\centering
\begin{tabular}{|l|l|}
\hline
\textbf{Notations} & \textbf{\makecell{Meaning}} \\ \hline
        $E$ & \makecell{Total global epochs of Phase-1} \\ \hline
        n & \makecell{Total clients in a global epoch} \\ \hline
        $E_i$ & \makecell{Epochs of $ith$ client in Phase-2}\\ \hline
        $M_i^{CF}$, $M_i^{S}$, $M_i^{CB}$  & \makecell{Front, Central model, and Back model\\ of the $ith$ client} \\ \hline
        $W_i^{CF}$, $W_i^{S}$, $W_i^{CB}$ & \makecell{Front, Central, and Back model's weights\\ of the $ith$ client} \\ \hline
        $d_i$ & \makecell{Number of training instances \\ with the $ith$ client}\\ \hline
        $A_{i,b}^{CF}$, $A_{i,b}^{S}$ & \makecell{Batch b activations of the Front \\ and central layers of $ith$ Client}\\ \hline
        $\eta$ & \makecell{Learning rate}\\ \hline
        $l_{i,b}$ & \makecell{Loss of batch b of $ith$ client}\\
        \hline
        $\frac{dl_{i,b}}{dA_{i,b}^S}$ & \makecell{Gradient of loss of batch b \\of $ith$ client with respect to $A_{i,b}^S$} \\ \hline
        $B$ & \makecell{Total Batches present at any client}\\ \hline
        $Y_{i,b}^{true}, Y_{i,b}^{pred}$ & \makecell{True and Predicted labels \\of the $ith$ client}\\ \hline
        $UFL$ & \makecell{Unfrozen active layers} \\ \hline  
        $V_{thres}$ & \makecell{Threshold validation accuracy}\\ \hline
        $O_{i}$ & \makecell{ Work done by client $i$}\\ \hline
        
\end{tabular}
\caption{Summary of Notations used in the paper, i in subscript refers to the $ith$ client}
\label{tab:notations}
\end{table} 

For ease of understanding, Algorithms 1a, 1b, and 1c describe the details of what happens during a round. Each round comprises two phases. Firstly, all clients in our framework enter the generalization phase (1) consisting of many global epochs. This phase trains all the clients in parallel to make sure that each of their models becomes generalized, i.e. it can perform the given task on unseen test data coming from a general data distribution with high performance. This stage is essential for achieving good personalized performance and not doing it may lead to over-fitting and poor performance even on the specific data distribution at the client because of the small number of training instances present at a client. After the model performance has reached convergence on the validation dataset, the personalized phase (2) is triggered. In this phase, each client focuses only on its own personal objective and customizes its model to perform well on its own data distribution. To summarize, in phase 1, clients work for the common good and in phase 2, they maximize personal good. 

The $ith$ client completes its local epoch by completing a forward and backward pass through the split layers and then updating its weights. This is followed by a call to the $ith$ offloading server instance to communicate this completion. The $ith$ client immediately calls the secure client model server with its client back layer weights. Both the averaging servers accept connections from their callers and wait for a timeout period, after which they average the weights of the connections and return them the averaged weight tensor. The $ith$ client waits for the confirmation of weight averaging from the offloading server before proceeding to its next local epoch.

In Phase 1, we run a fixed upper limit of E global epochs at all clients (to limit resource utilization on the client's side) or stop when a convergence criterion is reached. At this point, all the participating clients have trained a global generalized model. They are now given a choice to leave or participate in further training in the Personalization Phase (2).  In Phase 2, we freeze all the layers of $ith$ client's central layers at the offloading server and only train the back layers with the client. In this phase, there is no synchronization step of weight merging at the clients and the server instances. This allows all the clients to train their models for different numbers of epochs ($E_i$) than other clients and thus be completely independent of one another.  This Phase trains only the $ith$ client's back layers on its personal dataset, thus achieving high performance.

Weight averaging at the central and client back layers has been proposed previously \cite{split_fed}. We now describe some of the novel features of our algorithms in more detail. 
\begin{itemize}
    
    \item {Our algorithm is flexible to allow clients to leave in between the phases or even inside a phase which makes it highly suitable for resource-constrained devices imposing no constraint on them to remain in training for long. It also gracefully handles client disconnects and failures.}
    
    \item{The use of a validation set to determine important parameters of a distributed learning process although essential has not been clearly defined before. By using a validation set, we dynamically change the duration of our generalized phase. If the average validation accuracy of all clients exceeds a validation threshold, we can stop the generalization phase for all clients and let them decide to continue to the next phase or leave. }
    
    \item{The procedure CB defines the details of training back layers that are private to the clients. These are the most crucial layers for personalization.  }

\end{itemize}
To the best of our knowledge, we are the first to incorporate seamless, personalized learning in a distributed split learning framework by introducing the concept of phased learning, where we optimize varying objectives of participating clients, ensuring their best interests. Existing works \cite{split_comp_fed, split_fed, split_main} train clients end-to-end for the generalized objective only, ignoring any specific requirements of a particular client for personalized models. On the other hand, our lightweight personalization phase decouples a client from the other clients and lets them use our framework to better their model's performance on their personal dataset independently.

\begin{subalgorithms}

\begin{breakablealgorithm}
\caption{PFSL: Client Model Averaging Server}
\label{alg:sec_cms}

\begin{algorithmic}[1]

\Procedure{MergeWeights\_Clients}{$W_i$, $V_i$}

\noindent
\colorbox{pink}{\parbox{8cm}{
   \State \small{\textit{{/* Separate thread for each client connection */}}}
   \State Wait(t)
   \noindent

    \State \small{\textit{ /* Assuming m connected clients after time t */}}
    \State $ W_{avg} = (1/m) * \sum_{i=1}^{m} W_i $ 
    
    \For{$i=1$ to $m$} 
        \State $W_i = W_{avg}$
    \EndFor

    \State \small{\textit{ /* Avg. validation accuracy of connected clients*/}}

    \State $ V_{avg} = (1/m) * \sum_{i=1}^{m} V_i $ 

    \If{$V_{avg} \geq V_{thres}$}
        \State conv $=$ true
    \EndIf
    
}}    

    \State \textbf{Return $W_i$, conv}

\EndProcedure

\end{algorithmic}
\end{breakablealgorithm}

\begin{breakablealgorithm}
\caption{PFSL: Client}
\label{alg:PSFL_client}

\begin{algorithmic}[1]

\Procedure{Init\_Client}{i}
    \State Load pre-trained weights in $W_i^{CF}, W_i^{CB}$
    \State Freeze all layers of $M_i^{CF}$
\EndProcedure

\Procedure{CF}{i, phase}
    
    \For {$b=1$ to B}
       \State $A_{i,b}^{CF} = M_i^{CF}(b)$
       \State \small{\textit{/* Wait below for ith server instance to return */}}
        \State $val \gets offloading\_server.Forward(i,A_{i,b}^{CF}, phase)$ 
    \EndFor 
    \If{val}
        \State \textbf{Return} true
    \EndIf
\EndProcedure

\Procedure{CB}{i, $A_{i,b}^S$, phase}    

\noindent
\colorbox{pink}{\parbox{7cm}{    
   \State $Y_{i,b}^{pred} \gets M_i^{CB}(A_{i,b}^S)$ 
    \State Calculate $l_{i,b} = Loss(Y_{i,b}^{pred}, Y_{i,b}^{true})$
   \State Calculate $\frac{dl_{i,b}}{dW_i^{CB}}$ and $\frac{dl_{i,b}}{dA_{i,b}^S}$ 
   \State $W_i^{CB} = W_i^{CB} - \eta \frac{dl_{i,b}}{dW_i^{CB}}$
   \If{phase is 1}
        \State Calculate $\frac{dl_{i,b}}{dA_{i,b}^S}$
        \State Send  $\frac{dl_{i,b}}{dA_{i,b}^S}$ to the offloading Server
    \Else
        \State \textbf{Return} true
    \EndIf
}}

\EndProcedure

\Procedure{Client}{i}

    \State Init\_Client(i)
    \State Init\_offloading\_Server(i)
     \State *** \textbf{Generalization Phase (1)} ***
    \State Set phase = 1
    \While{(e $=$ E) or conv}
        \State val2 $\gets$ CF(i,phase)  \Comment{Wait for val2 $=$ true}
        \State \small{\textit{/* Calculate validation set accuracy on a common validation set V*/}}
        \State $V_i$ = val\_acc(V)
        \If{val2}
            \State offloading\_server.Server(i)  
            \State $W_i^{CB}, conv = $MergeWeights\_Clients($W_i^{CB}$, $V_i$) 
            \State val $\gets$ offloading\_server.Server(i, conv) \Comment{Wait here} \label{lst:line: server_complete}
            \If{val}
                \State Proceed
            \EndIf 
        \EndIf
    \EndWhile

    \State \small{\textit{/* Client i can leave now or continue */}}

\noindent
\colorbox{pink}{\parbox{8cm}{    
    \State *** \textbf{Personalization Phase (2)} ***
    \State Set phase = 2

    \For {$e=1$ to $E_i$} 
        \For{$i=1$ to $n$}
            \State val $\gets$ CF(i, phase) \Comment{Wait here}
            \If{val}
                \State continue
            \EndIf
        \EndFor
    \EndFor 
}}

\EndProcedure

\end{algorithmic}
\end{breakablealgorithm}

\hfill\vline\hfill

\begin{breakablealgorithm}
\caption{PFSL: Offloading Server}
\label{alg:PSFL_server}
\begin{algorithmic}[1]

\Procedure{MergeWeights}{$W_i$}

    \State \small{\textit{/* Wait below for time $t_1$ before proceeding */}}
    \State Wait($t_1$)
    \State \small{\textit{/* Assuming $k_1$ connected clients after time t1 */}}
    \State $ W_{avg} = (1/k_1) * \sum_{i=1}^{k_1} W_i $ 
    \For{$i=1$ to $k_1$} 
        \State $W_i = W_{avg}$
    \EndFor

    \State \textbf{Return $W_i$}

\EndProcedure

\Procedure{Init\_Offloading\_Server}{i}
        \State Load pre-trained weights in $W_i^{S}$
        \State Freeze k layers from the start of $M_i^S$
\EndProcedure

\Procedure{Forward}{i,$A_{i,b}^{CF}$, phase}

   \State $A_{i,b}^S = M_{i}^{S}(A_{i,b}^{CF})$
    \If{phase is 1}  
        \State \small{\textit{/* Wait below for CB of ith client to return */}}
        \State $\frac{dl_{i,b}}{dA_{i,b}^S} \gets client.CB(i,A_{i,b}^S, phase)$   
        \State $ \frac{dl_{i,b}}{dW_{i,UFL}^S}  \gets $ Backprop($\frac{dl_{i,b}}{dA_{i,b}^S}$)  
        \State $W_{i,UFL}^S = W_{i,UFL}^S - \eta \frac{dl_{i,b}}{dW_{i,UFL}^S}$ 
        \State \textbf{Return} true
    \Else
        \State \small{\textit{/* Wait below for ith Client Back to return true */}}
        \State $val = client.CB(i,A_{i,b}^S, phase)$ 
        \If{val}
            \State \textbf{Return} true
        \EndIf
    \EndIf
\EndProcedure

\Procedure{Server}{i}

    \While{ (e $=$ E) or conv }
        \State \small{\textit{/* Wait below for request from the client */}}
        \State receive()
        \State $W_i^{S} = $MergeWeights($W_i^{S}$)
        \State \small{\textit{/* Wait below for request from the client */}}
        \State conv $=$ receive(client, conv)
        \State \textbf{Return} True to client
    \EndWhile

    \State Freeze all layers of $M_i^{S}$
\EndProcedure

\end{algorithmic}
\end{breakablealgorithm}

\end{subalgorithms}

%% file: sections/Experiments_and_results.tex
\section{Experiments and Results}
\label{sec:experiments_and_results}
We have implemented our framework, PFSL, in PyTorch. Our implementation is modular and allows specifying the number of clients, DNN model, number of the front, central and back layers, number of data points per client, etc. We have implemented state-of-the-art DML algorithms: FL, FL\_TL \cite{fl_tl}, SL \cite{split_main}, SFLv1, and SFLv2 \cite{split_fed} as described in their papers and reproduced their results for the validating our implementations. For experimentation, we used our workstation with Intel(R) Xeon(R) Gold 5218 CPU @ 2.30GHz, with NVIDIA RTX A6000 GPU.

In all our experiments, we use ResNet-18 \cite{resnet18 } with pre-trained weights, which provides a good balance between accuracy and compute resource requirements. We use 2 ophthalmic imaging datasets for blindness detection: APTOS\cite{aptos2019blindness} and EyePACS \cite{kaggleKaggleDataset} and normalize image size to 224x224. We have used several image classification benchmarks: MNIST with 10 classes of hand-written digits \cite{MNIST},  F-MNIST\cite{fmnist} with 10 classes of fashion products, and CIFAR-10\cite{cifar-100} with 10 classes of objects. The size of input images in these benchmarks is 32x32. Since ResNet-18 was pre-trained with 224x224 size and weights of the initial layers were frozen for all algorithms with transfer learning (PFSL, FL\_TL), these images had to be scaled to 224x224 for them to achieve good performance. 

In the real world, good quality labeled data is a scarce resource. The clients are heterogeneous and their data distributions of clients are often non-i.i.d. Similar to Federated Learning, PFSL is applicable for diverse applications. Thus, we carefully evaluate performance and fairness in practical scenarios/settings. We consider the following settings:
\begin{itemize}
\item{S1: Labels are uniformly distributed, number of labeled data points per client is the same but small.}
\item{S2: Labels are non-uniformly distributed, the number of labeled data points per client is the same but small.}
\item{S3: Labels are uniformly distributed. One client has a high number of labeled data points and the remaining have small.}
\item{S4: All clients have a high number of labeled data points for MNIST, F-MNIST, and CIFAR-10.}
\item{S5: 1000 clients in the system each participating only for 1 epoch, and 50\% of them dropping out before completing training.}
\item{S6: Clients have data coming from different benchmarks: APTOS and EyePACS.}
\end{itemize}

For scenarios S1, S2, S3, and S5, we use CIFAR-10 as it is quite complex, and achieving good accuracy is non-trivial. Apart from the existing DML algorithms, for performance comparison, we also include a centralized model assuming that it has access to the labeled data from all the clients. It is expected to have the highest performance but can compromise data privacy. Hence, the goal of DML algorithms is to equal their performance without compromising accuracy. We have two versions of Central models (with and without TL) to compare with the performance of relevant algorithms. To quantify the benefit of DML, we also compare it with the client's performance by performing training locally without participating in DML. We call this individual training. For individual clients, starting from pre-trained weights is best, so only that is reported (as Individual\_TL). 

\begin{figure}[H]
\includegraphics[width=\linewidth]{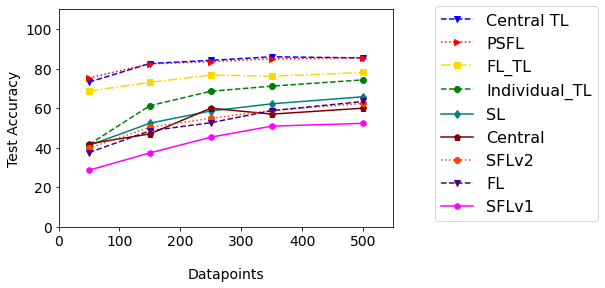}
\caption{S1: Test Accuracy for DML algorithms vs Datapoints} 
\label{Test_setting1}
\end{figure}
\vspace{-1 em}

\subsection{Setting 1: Small Sample Size (Equal), i.i.d.}
Each client has a very small number of labeled data points, and all these samples are distributed identically across clients. The number of data points per client is slowly increased to study the benefits of having more data per client on the global model's accuracy. For each algorithm, we keep the global epochs fixed at 100 and the number of participating clients at 10. We evaluate each client on a CIFAR-10 test set having 2000 data points with the same distribution and report the average test accuracy. Since the test and train data distributions across all clients are the same, we run our algorithm only till Generalization Phase 1. For FL\_TL, the same number of layers were unfrozen at the client's side as PFSL for a fair comparison.

There are several interesting observations from Fig \ref{Test_setting1}. PFSL achieves accuracy close to the Central model (with TL) with the complete dataset. Its accuracy exceeds that of FL\_TL by 10\%. It is clear from this experiment, that for realistic settings with small sample size, transfer learning is very helpful, since Individual\_TL also exceeds the accuracy of FL, SL, SFLv1, and SFLv2 which attempt to train from scratch.

\begin{figure}[H]
\centering
\includegraphics[width=0.6\linewidth]{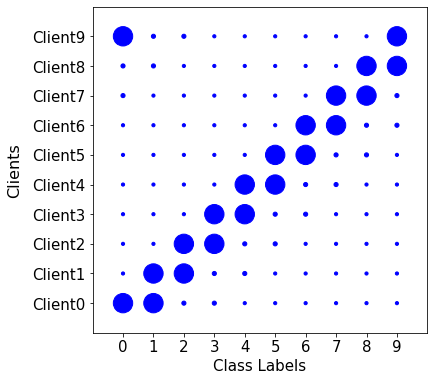}
\caption{Data Distribution among clients where larger circle indicates more number of training samples for that particular class and client  } 
\label{Setting1_distribution_plot}
\end{figure}

\subsection{Setting 2: Small Sample Size (Equal), non-i.i.d.}

In this scenario, we consider the clients to have a small sample size and different data distributions, as shown in Fig \ref{Setting1_distribution_plot}. There are a total of 10 clients, and each client has 500 labeled data points. We model a situation where every client has more labeled data points from a subset of classes \emph{(prominent classes)} and less from the remaining classes. We chose to experiment with heavy label imbalance and diversity. The test distribution of each client is the same as its train distribution providing an incentive for them to run \textbf{Personalization Phase 2} as well. To evaluate algorithms in this setting, we consider the average F1 scores of the \emph{prominent classes} for each client on its respective test set and then average these over all the clients. As the data distribution across all clients differs, we run our algorithm completely until Phase 2. We run Phase 1 till convergence and Phase 2 again till convergence. 

\begin{table}[H]
    \centering
    \begin{tabular}{|l|l|l|l|}
    \hline
        \textbf{Algorithm} & \textbf{F1 score} & \textbf{Epochs} & \textbf{Test Standard Deviation}\\ \hline
        FL & 0.505 & 100 & 8\\ \hline
        SL & 0.794 & 50 & 4.25\\ \hline
        SFLv1 & 0.749 & 50 & 5\\ \hline
        SFLv2 & 0.773 & 50 & 4.25\\ \hline
        PFSL$_{BL\_3} (gen)$ & \textbf{0.816}  & \textbf{24} & \textbf{5.25}\\ \hline
        PFSL$_{BL\_3} (pers)$ & \textbf{0.953} & \textbf{26} & \textbf{2}\\ \hline
        PFSL$_{BL\_2} (gen)$ & \textbf{0.79}  & \textbf{24} & \textbf{5.75}\\ \hline
        PFSL$_{BL\_2} (pers)$ & \textbf{0.95} & \textbf{30} & \textbf{2.1}\\ \hline
        \end{tabular}
    \caption{S2: Average F1 Scores of 10 clients}
    \label{Setting1: f1 table}
\end{table}

From Table \ref{Setting1: f1 table}, we can observe that the average F1 score of Split learning and its variants is much better than Federated learning. Our algorithm, PFSL, gives the best average F1 scores. This happens because phase 2 allows the clients to train unique models suited to their own train datasets. Moreover, the generalization phase of our algorithm ends at the 25th epoch with better average F1 scores than the other algorithms achieve at 100 epochs. Our lightweight personalization phase results in a significant boost of 0.14 in average F1 scores. PFSL$_{BL\_3}$ has used three back layers at the client while  PFSL$_{BL\_2}$ used only two back layers at the client, which is our default. We observe that increasing the number of layers at the back model only affects the number of epochs needed to converge as the F1 scores converge to the same value in both cases. We notice that once the clients have converged to a generalized model in our framework, its personalization is very quick (2 global epochs in $PFSL_{BL\_3}$ and 6 global epochs in $PFSL_{BL\_2}$). 

The Test Accuracy Standard deviation column shows the values of Performance Fairness of different DML algorithms using the standard deviation of test performance of the clients in different frameworks. We can observe from Table \ref{Setting1: f1 table} that PSFL shows comparable performance fairness to others in the generalization phase. But after running the personalization phase, the fairness increases dramatically, indicating the ability of all the clients to converge easily on their respective personalized models, given the starting generalized weights. This also shows that the weights obtained from the generalization phase need to be fine-tuned to fit individual data better, as described in \cite{ditto_main, P1, P2, P3, P4} to improve fairness. We believe that in a practical setting, the ability to personalize private models is very attractive, and the performance gain and high-performance fairness provide an incentive to clients to work together for the common good while still enjoying the competitive edge.

\subsection{Setting 3: Small Sample Size (Unequal), i.i.d.}
Very often, the participating clients may have a different sample size. Most clients usually have fewer data points, and a select few have a much larger dataset. In this situation, the clients with more data tend to think that participation in distributed training with the clients (who may have fewer data) won't greatly benefit them, and they may lose their competitive edge. Thus, they can decide to forego participating in training completely.

To simulate the above-described situation, we consider 11 clients where the \emph{Large client} has 2000 labeled data points while the other ten small clients have 150 labeled data points, each distributed identically. Note that the class distributions among all the clients are the same. For evaluation purposes, we consider a test set having 2000 data points with an identical distribution of classes as the train set. Firstly, we calculate the average test accuracy when only 10 small clients $C_1 - C_{10}$ with 150 data points each participate. In the next case, all 11 clients (small and Large) participate in training, and we calculate the first client's test accuracy individually and the remaining clients' average test accuracy. We also calculate the performance achieved by the \emph{Large client} if it begins transfer learning on a pre-trained model utilizing only its 2000 labeled data points. To give incentive to the Large client, its test accuracy must increase significantly by participating in the learning with other clients with fewer data.

Existing works have not focused on achieving work fairness when the number of training data points per client is imbalanced. We consider that it is unfair to make the \emph{Large client} work ten times more than each small client if it has ten times more data than them. As we have explained before, in PFSL, we ensure that the number of batches and the batch size with each client in a global epoch remain the same. We calculate the metrics of this setting on our algorithm after ensuring this fairness constraint. Also, since the data distributions across all clients are the same, we run our algorithm only till Phase 1.

We present the Average test and train performance when only the small clients participate in DML training in Table \ref{setting2_t1}. It is evident that PFSL achieves the highest accuracy and minimizes over-fitting. Next, we present the Average test performance of small and \emph{Large client} when all the 11 clients participate in DML training in Table \ref{setting2_t2}. We can see a significant increment in their test accuracies in all the algorithms. This shows that the small clients greatly benefit from the \emph{Large client}. 

If \emph{Large client} is only trained on its dataset of 2000 data points, using a pre-trained model and transfer learning, then the test accuracy is 83.  Table \ref{setting2_t2} shows that only PFSL can give an incentive to such clients by providing a 2\% increase in accuracy. Moreover, our results are calculated after incorporating work fairness. We do not force the \emph{Large client} with more data to do more work in a local epoch than the other smaller clients, yet provide tangible benefits and incentives to participate in DML. We observe that when we make the \textit{Large client} do more work, its accuracy does not increase significantly (2\% over the previous) but its work done per global epoch increases by 16 times. Thus, by work fairness, we ensure high performance for this client with a far lesser workload.  

\begin{table}
\setlength{\tabcolsep}{2.5pt}
\centering
    \begin{tabular}{|l|l|l|}
    \hline
        \textbf{Algorithm} & \textbf{Avg $C_1 - C_{10}$ Test} & \textbf{Avg $C_1 - C_{10}$ Train} \\ \hline
        FL & 48.56 & 86.25 \\ \hline
        SL & 52.42 & 88.01 \\ \hline
        SFLv2 & 50.43 & 85.35 \\ \hline
        SFLv1 & 37.37 & 81.23 \\ \hline
        \textbf{PFSL} & \textbf{81.42} & 100 \\ \hline
    \end{tabular}
    \caption{ S3: Test and Train accuracies when only  $C_1 - C_{10}$ participate in DML}
    \label{setting2_t1}
    \bigskip 
\begin{tabular}{|l|l|l|}
    \hline
        \textbf{Algorithm} & \textbf{Avg $C_1 - C_{10}$ Test } & \textbf{\emph{Large Client} Test} \\ \hline
        FL & 59.02  & 59.15 \\ \hline
        SL & 63.41  & 64.64 \\ \hline
        SFLv2 & 61.01  & 61.96 \\ \hline
        SFLv1 & 56.76 & 62.12 \\ \hline
        \textbf{PFSL} & \textbf{84.62}  & \textbf{84.64} \\ \hline
    \end{tabular}
    \caption{S3: Test Accuracies of Clients $C_1 - C_{10}$ and \emph{Large Client} when all participate in DML}
    \label{setting2_t2}
    
\end{table}

\subsection{Setting 4: A large number of data samples}
\begin{figure}[H]
\includegraphics[width= \linewidth]{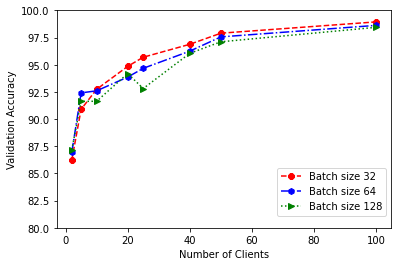}
\caption{S4: Hyper-parameter tuning using the validation set for MNIST} 
\label{val_plots}
\end{figure}

For the sake of comparison with most existing works, where a large number of samples per client are assumed, we experimented with three different image classification datasets: MNIST, FMNIST, and CIFAR-10. We also tune the hyperparameters of our algorithm on the validation set of the MNIST dataset as shown in Fig. \ref{val_plots}. This helps in giving an estimate of the parameters that work best for achieving optimal performance of all clients in our framework. Five clients were used in the experiment and each client had 10k datapoints for CIFAR-10, 12k MNIST and 12k for FMNIST. These clients then participate in distributed learning and try to achieve performance close to centralized learning. We evaluate all the clients by their accuracies on a common test set. We report the averaged train and test accuracies for each algorithm. 

Table \ref{tab:pop_bm_1} reports the average test accuracies of the participating clients across all the algorithms on CIFAR-10, FMNIST, and MNIST datasets when each of them has a large number of training data points. Our algorithm gives the best performance on all the datasets. Notably, it performs significantly better than the other algorithms on the CIFAR-10 dataset, which is more complex than the other two. 

Table \ref{tab : pop_bm_2} shows the average train and test accuracies of 5 clients and the amount of overfitting (difference between train and test accuracies) across all algorithms on the CIFAR-10 dataset. We can observe that other algorithms heavily overfit their training data and thus perform poorly on their test data. In contrast, our algorithm shows the least amount of over-fitting because of a carefully designed weight-averaging process. 

\begin{table}
\setlength{\tabcolsep}{2.5pt}
\centering
\begin{tabular}{|l|l|l|l|}
\hline
\textbf{Algorithm} & \textbf{CIFAR-10} & \textbf{FMNIST} & \textbf{MNIST} \\ \hline
$d_i$ & 10K & 12K & 12K \\ \hline
FL  & 81.52 & 91.38  & 99.26 \\ \hline
SL & 82 & 91.08 & 99.14 \\ \hline
SFLv1 \cite{split_fed}  & 78.82 & 90.55  & 98.7\\ \hline
SFLv2 \cite{split_fed} & 73.82 & 90.93 & 99.08 \\ \hline
\textbf{PFSL} & \textbf{92} & \textbf{93} & \textbf{99.46}  \\ \hline
\end{tabular}
\caption{S4: Average Test Accuracy, five clients on three important image classification benchmarks}
\label{tab:pop_bm_1}
\bigskip 
\begin{tabular}{|l|l|l|l|}
    \hline
        \textbf{Algorithm} & \textbf{Train Accuracy} & \textbf{Test Accuracy} & \textbf{Overfitting} \\ \hline
        FL & 92 & 81.52 & 10.48 \\ \hline
        SL & 96.94 & 82 & 14.94 \\ \hline
        SFLv1 & 92.73 & 78.82 & 13.91 \\ \hline
        SFLv2 & 96.5 & 73.82 & 22.68 \\ \hline
        \textbf{PFSL} & \textbf{99.8} & \textbf{92} & \textbf{7.8} \\ \hline
    \end{tabular}
    \caption{S4: Overfitting on CIFAR-10, 5 clients}
    \label{tab : pop_bm_2}
  
\end{table}

\subsection{Setting 5: System simulation with 1000 clients}

\begin{figure}[H]
\includegraphics[width=\linewidth]{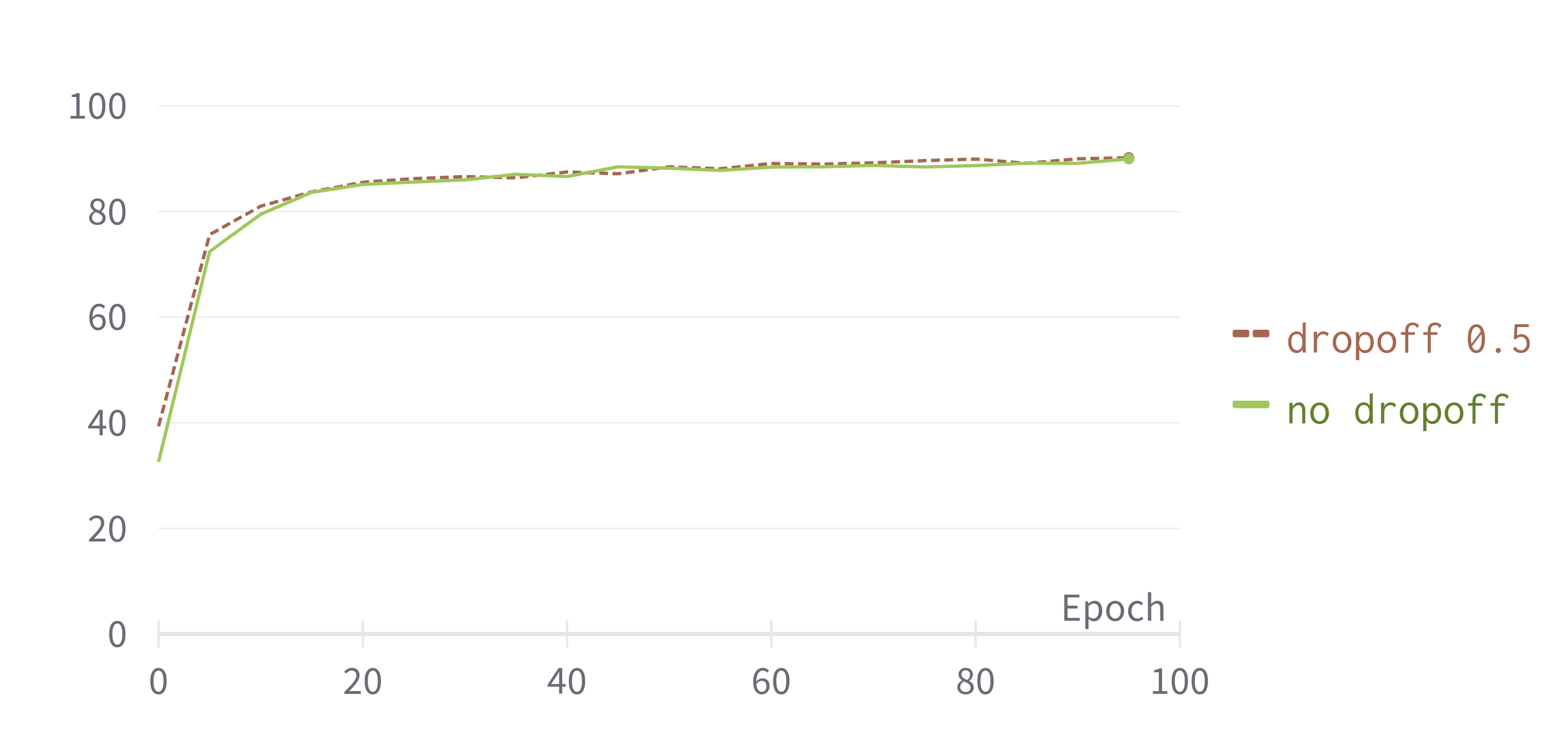}
\caption{S5: Test Accuracy with number of epochs} 
\label{fig: sys_sim}
\end{figure}

In this setting, we simulate 1000 clients.  We allow only 10 clients to simultaneously perform the training. Each client stays in the system only for 1 round which lasts only 1 epoch. Thus, we evaluate our system for the worst possible scenario when every client cannot stay in the system for long and can only afford to make a minimal effort to participate. We assume that each client has 50 labeled data points sampled randomly but unique to the client. Within each round, we simulate a dropout, where clients begin training but are not able to complete the weight averaging. We keep the dropout probability at 50\%. In this experiment, we used the CIFAR-10 dataset. We find that the final test accuracy after 100 rounds (where each client begins the training exactly once), was 90\% for PFSL (see Fig.\ref{fig: sys_sim}). This is very close to that observed in an ideal scenario with 5 clients having 10K data points each (see Table \ref{tab : pop_bm_2}). This setting clearly demonstrates the high scalability of our framework in realistic scenarios with extremely resource-constrained devices participating in the distributed learning process.

\subsection{Setting 6: Different Diabetic Retinopathy Datasets}
This experiment describes the realistic scenario when healthcare centers have different sets of raw patient data for the same disease. We have used two datasets EyePACS \cite{eyePacs} and APTOS\cite{aptos2019blindness}. The images in EyePACS were captured under various conditions by various devices at multiple primary care sites throughout California and elsewhere. The APTOS dataset, on the other hand, has been collected across various healthcare sites in India. For the experiment, we created a set-up where there was a total of 10 clients, in which the first 5 clients were provided the APTOS dataset and the next 5 clients (Client 5- Client 9) were provided the EyePACS dataset. The sample size of data for each client was kept at 500. The images had to be classified into three categories-No DR(0), Moderate(1), and Severe(2). All clients ran the personalization phase to maximize their test accuracies on their respective datasets.

Table \ref{tab:algo_f1_score} shows the average test accuracies achieved by the first set of clients and the second set of clients separately. We have also shown the F1 Scores of one representative client having the APTOS dataset and one of the clients having the EyePACS dataset across all the algorithms.
\begin{table}[H]
\centering
\begin{tabular}{|l|l|l|l|l|l|}
\hline
\textbf{Algorithm} &  \textbf{$C_0$ F1 } &\textbf{$C_5$ F1 } & \textbf{$C_0-C_4$ Test}  & \textbf{$C_5-C_9$ Test}  \\ \hline

FL  & 0.73 & 0.36	 & 81.7	& 70.14 \\ \hline
SL & 0.72 & 0.41 & 80.37 &	69.57 \\ \hline
SFLv1 \cite{split_fed}  & 0.61 & 0.32 &	77.79 &	69.55\\ \hline
SFLv2 \cite{split_fed} & 0.74 & 0.43 & 80.21 &	66.53  \\ \hline
\textbf{PFSL} & \textbf{0.78}  & \textbf{0.58} & \textbf{85.36} & \textbf{70.86}  \\ \hline

\end{tabular}
\caption{S6: F1 Score of Client 0 (having APTOS dataset) and F1 Score of Client 5 (having EyePACS dataset) and Avg Test Accuracies of Client 0-4 (APTOS dataset) and Client 5-9 (EyePACS dataset)}
\label{tab:algo_f1_score}
\end{table}

From the results, we see that even though the two datasets are very different, the PFSL algorithm achieves the highest F1 scores for Client 0 and Client 5 and also the highest average test accuracies for the first and second sets of clients. This shows that our algorithm achieves the best, personalized results, wherein each set of clients performs very well on their own dataset.

\subsection{Work Fairness Analysis}

Table \ref{tab : wfair} shows the work done by each client in different DML algorithms. As we can see, there is a difference in values of $C_i$ for S3 and S6 for FL, SL, SFLv1, SFLv2, because of the change in the size of input images from 32x32 in S3 to 224x224 in S6. For PFSL, we had to use 224x224 in both S3 and S6 because of using transfer learning on ResNet-18 pre-trained model. This could have been reduced considerably by using a customized pre-trained model for 32x32. It is evident that FL algorithms are very resource intensive as compared to all the SL variants. In S3, we can see the clear benefit of work fairness for the large client (O\_L\_c) under PFSL. All other DML algorithms cause the large client to overwork heavily for marginal gains. As we mentioned, a better approach would be to let the large client participate in multiple rounds to increase its accuracy gradually. Finally, in Setting 6, we can see the clear benefits of using PFSL over all other DML algorithms.

\begin{table}[H]
    \centering
    \begin{tabular}{|p{1.5cm}|l|l|p{1cm}|p{0.8cm}|l|}
    \hline
        \textbf{Algorithm } & \textbf{S3:C\_i} & \textbf{S6:C\_i} & \textbf{S3: O\_L\_c} & \textbf{S3: O\_S\_c } & \textbf{S6: O\_i} \\ \hline
        FL  & 14.28 & 700.2 &44268 & 2856 & 105030 \\ \hline
        SL, SFLv1, SFLv2 & 2.76 &  135.9 & 8556& 552 & 10192 \\ \hline
        PFSL  & 15.5 & 15.5 & 775& 775 & 744 \\ \hline
    \end{tabular}
    \caption{Work Fairness Analysis for settings S3 and S6. C refers to the work done per iteration and O refers to the total work done till convergence is reached. L\_c refers to the Large client in S3 and S\_c refers to the small clients in S3. All values are computed in GFlops using a profiler \cite{profiler}}
    \label{tab : wfair}
\end{table}

%% file: sections/Conclusion.tex
\section{Conclusion}
\label{sec:conclusion}
We have implemented, PFSL, by incorporating transfer learning with lightweight personalization and work fairness while ensuring privacy for both input data and labels. Our extensive empirical studies have demonstrated that it is computationally efficient for thin clients and outperforms FL-TL and current 2-stage SL variants in accuracy and fairness metrics. We have shown the advantages of personalization for split learning for clients with non-i.i.d. data distributions. Our fair work balance constraint doesn't add any additional burden on clients with the larger quantum of data while achieving performance gains. Our work suggests several interesting directions for future study including proving PFSL's convergence and privacy guarantees theoretically, studying its performance under adversarial attacks, hardware implementation and further optimization of resource consumption overheads.

%% file: sections/Appendix.tex
 \clearpage
 \appendices

  \subsection{Artifact Appendix}
  \label{appendix:artifact}
    Software is being developed and maintained here: \url{https://github.com/mnswdhw/PFSL}

    A persistent identifier for the software code can be found here: \url{https://doi.org/10.5281/zenodo.7739655 }
  
  We have tested the following commands in Ubuntu 20.04.5 LTS. 
  
  \subsubsection{Build requirements}
  \label{appendix:build-req}
  \begin{itemize}
      \item Python3 ($>$=3.8)
      \item pip 21.0.1
      \item Nvidia GPU ($>$=12GB)
      \item Conda 
  \end{itemize}
 
 You may use the following steps to install the required libraries:
 \begin{itemize}
     \item Change Directory into the project folder
     \item Create a conda environment using the command \textit{conda create \texttt{-{}-}name {env\_name} python=3.8} Eg- conda create \texttt{-{}-}name pfsl python=3.8
\item Activate conda environment using the command \textit{conda activate {env\_name}} Eg- conda activate pfsl
     \item pip install -r requirements.txt
    \item{Create a results directory in the project folder to store all
the resulting plots using the commands below:}
    \begin{itemize}
        \item mkdir results
        \item mkdir results/FL
        \item mkdir results/SL
        \item mkdir results/SFLv1
        \item mkdir results/SFLv2
    \end{itemize}
 \end{itemize}
 
\subsection{Algorithm Description}
  \label{appendix:testrun}

  The main implementaion of the PFSL algorithm, algorithm 1b and 1c is in PFSL.py from where all the function calls for client and server are made.

\begin{itemize}
    \item Client class is implemented in client.py. All the function calls for each client is defined here.
    \item Server class is implemented in ConnectedClient.py. All the function calls for the server copy of each client is defined here.
    \item  Algorithm 1c for merging of weights is implemented in utils/merge.py
\end{itemize}

  \subsection{Test Run}
  \label{appendix:testrun}

The parameters options for a particular file can be checked adding --help argument.
 Optional arguments available for PFSL are:

 \begin{itemize}
    
 \item  -h, - -help            show this help message and exit 
 \item  -c , \texttt{-{}-}number\_of\_clients  Number of Clients (default: 10)
 \item  -b , \texttt{-{}-}batch\_size   Batch size (default: 128)
 \item  \texttt{-{}-}test\_batch\_size   Input batch size for testing (default: 128)
 \item  -n , \texttt{-{}-}epochs      Total number of epochs to train (default: 10)
 \item  \texttt{-{}-}lr                Learning rate (default: 0.001)
 \item  \texttt{-{}-}dataset      States dataset to be used (default: cifar10)
 \item  \texttt{-{}-}seed            Random seed (default: 1234)
 \item  \texttt{-{}-}model          Model you would like to train (default: resnet18)
 \item  \texttt{-{}-}opt\_iden   optional identifier of experiment 
 \item  \texttt{-{}-}pretrained          Use transfer learning using a pretrained model (default: False)
 \item  \texttt{-{}-}datapoints  Number of samples of training data allotted to each client (default: 500)
 \item  \texttt{-{}-}setting     Setting you would like to run for, i.e, setting1 , setting2 or setting4 (default: setting1)
 \item  \texttt{-{}-}checkpoint  Epoch at which personalisation phase will start (default: 50)
  \item  \texttt{-{}-}rate This arguments specifies the fraction of clients dropped off in every epoch (used in setting 5)(default: 0.5)
  
 \end{itemize}

 For reproducing the results, always add argument \texttt{-{}-}pretrained  while running the PFSL script.
 
  \subsubsection{\textbf{Setting 1: Small Sample Size (Equal), i.i.d.}}
  \label{appendix_setting1}
  To run all the algorithms for setting 1 argument  \texttt{-{}-}setting setting1 and \texttt{-{}-}datapoints [number of sample per client] has to be added. Rest of the arguments can be selected as per choice. Number of data samples can be chosen from 50, 150, 250, 350 and 500 to reproduce the results. When the datapoints per client was 50, batch size was chosen to be 32 and for other data samples greater than 50 batch size was kept at 64. Test batch size was always taken to be 512. For data sample 150, command are given below.
\begin{itemize}
      
  \item python PFSL\_Setting124.py  \texttt{-{}-}dataset cifar10 \texttt{-{}-}setting setting1 \texttt{-{}-}datapoints 150 \texttt{-{}-}pretrained \texttt{-{}-}model resnet18 -c 10 \texttt{-{}-}batch\_size 64 \texttt{-{}-}test\_batch\_size 512 \texttt{-{}-}epochs 100

  \item python FL.py     \texttt{-{}-}dataset cifar10 \texttt{-{}-}setting setting1 \texttt{-{}-}datapoints 150  -c 10 \texttt{-{}-}batch\_size 64 \texttt{-{}-}test\_batch\_size 512 \texttt{-{}-}epochs 100
  \item python SL.py     \texttt{-{}-}dataset cifar10 \texttt{-{}-}setting setting1 \texttt{-{}-}datapoints 150   -c 10 \texttt{-{}-}batch\_size 64 \texttt{-{}-}test\_batch\_size 512 \texttt{-{}-}epochs 100
  \item python SFLv1.py  \texttt{-{}-}dataset cifar10 \texttt{-{}-}setting setting1 \texttt{-{}-}datapoints 150  -c 10 \texttt{-{}-}batch\_size 64 \texttt{-{}-}test\_batch\_size 512 \texttt{-{}-}epochs 100
   \item python SFLv2.py \texttt{-{}-}dataset cifar10 \texttt{-{}-}setting setting1 \texttt{-{}-}datapoints 150 -c 10 \texttt{-{}-}batch\_size 64 \texttt{-{}-}test\_batch\_size 512 \texttt{-{}-}epochs 100
   \end{itemize}

  \subsubsection{ \textbf{Setting 2: Small Sample Size (Equal), non-i.i.d.}}

   To run all the algorithms for setting 2 argument  --setting setting2 has to be added. For PFSL, to enable personalisation phase from xth epoch, argument --checkpoint [x] has to be added. Rest of the arguments can be selected as per choice. 
\begin{itemize}
      
  \item python PFSL\_Setting124.py \texttt{-{}-}dataset cifar10 \texttt{-{}-}model resnet18 \texttt{-{}-}pretrained \texttt{-{}-}setting setting2 \texttt{-{}-}batch\_size 64 \texttt{-{}-}test\_batch\_size 512 \texttt{-{}-}checkpoint 25 \texttt{-{}-}epochs 30

  \item python FL.py \texttt{-{}-}dataset cifar10 \texttt{-{}-}setting setting2 -c 10 \texttt{-{}-}batch\_size 64 \texttt{-{}-}test\_batch\_size 512 \texttt{-{}-}epochs 100
  \item python SL.py \texttt{-{}-}dataset cifar10 \texttt{-{}-}setting setting2 -c 10 \texttt{-{}-}batch\_size 64 \texttt{-{}-}test\_batch\_size 512 \texttt{-{}-}epochs 100
  \item python SFLv1.py \texttt{-{}-}dataset cifar10 \texttt{-{}-}setting setting2 -c 10 \texttt{-{}-}batch\_size 64 \texttt{-{}-}test\_batch\_size 512 \texttt{-{}-}epochs 100 
   \item python SFLv2.py \texttt{-{}-}dataset cifar10 \texttt{-{}-}setting setting2 -c 10 \texttt{-{}-}batch\_size 64 \texttt{-{}-}test\_batch\_size 512 \texttt{-{}-}epochs 100
   \end{itemize}
  \label{appendix_setting2}

  \subsubsection{\textbf{Setting 3: Small Sample Size (Unequal), i.i.d.}}
In this setting, we consider we have 11 clients where the Large client has 2000 labelled data points while the other ten small clients have 150 labelled data points, each distributed identically. The class distributions among all the clients are the same. For evaluation purposes, we consider a test set having 2000 data points with an identical distribution of classes as the train set.

To reproduce Table IV of the paper, run setting 1 with datapoints as 150 as illustrated above. To reproduce Table V of the paper follow the below commands. In all the commands argument \texttt{-{}-}datapoints that denotes the number of datapoints of the large client has to be added. In our case it was 2000.
  \begin{itemize}
    
  \item python PFSL\_Setting3.py \texttt{-{}-}datapoints 2000 \texttt{-{}-}dataset cifar10 \texttt{-{}-}pretrained \texttt{-{}-}model resnet18 -c 11 \texttt{-{}-}epochs 50
  \item python FL\_Setting3.py \texttt{-{}-}datapoints 2000    \texttt{-{}-}dataset cifar10\_setting3  -c 11  \texttt{-{}-}epochs 100
  \item python SL\_Setting3.py  \texttt{-{}-}datapoints 2000   \texttt{-{}-}dataset cifar10\_setting3  -c 11  \texttt{-{}-}epochs 100
  \item python SFLv1\_Setting3.py \texttt{-{}-}datapoints 2000 \texttt{-{}-}dataset cifar10\_setting3  -c 11  \texttt{-{}-}epochs 100
   \item python SFLv2\_Setting3.py \texttt{-{}-}datapoints 2000 \texttt{-{}-}dataset cifar10\_setting3  -c 11  \texttt{-{}-}epochs 100
   \end{itemize}
  \label{appendix_Setting3}

\subsubsection{ \textbf{Setting 4: A large number of data samples}}
\label{appendix_setting4}

   To run all the algorithms for setting 4 argument  \texttt{-{}-}setting setting4 has to be added.The rest of the arguments can be selected as per choice. Dataset argument has 3 options: cifar10, mnist and fmnist.
\begin{itemize}
      
  \item python PFSL\_Setting124.py  \texttt{-{}-}dataset cifar10 \texttt{-{}-}setting setting4 \texttt{-{}-}pretrained \texttt{-{}-}model resnet18 -c 5 \texttt{-{}-}epochs 20

  \item python FL.py     \texttt{-{}-}dataset cifar10  \texttt{-{}-}setting setting4 -c 5 \texttt{-{}-}epochs 20
  \item python SL.py     \texttt{-{}-}dataset cifar10  \texttt{-{}-}setting setting4 -c 5 \texttt{-{}-}epochs 20
  \item python SFLv1.py  \texttt{-{}-}dataset cifar10  \texttt{-{}-}setting setting4 -c 5 \texttt{-{}-}epochs 20
   \item python SFLv2.py \texttt{-{}-}dataset cifar10  \texttt{-{}-}setting setting4 -c 5 \texttt{-{}-}epochs 20
   \end{itemize}

  \subsubsection{ \textbf{Setting 5: System simulation with 1000 client}}
\label{appendix_setting5}

In this setting we try to simulate an environment with 1000 clients. Each client stays in the system only for 1 round which lasts only 1 epoch. Thus, we evaluate our system for the worst possible scenario when every client cannot stay in the system for long and can only afford to make a minimal effort to participate. We assume that each client has 50 labeled data points sampled randomly but unique to the client. Within each round, we simulate a dropout, where clients begin training but are not able to complete the weight averaging. We keep the dropout probability at 50

Use the following command to reproduce the results:
\textit{Here rate argument specifies the dropoff rate which is the number of clients that will be dropped randomly in every epoch}

\begin{itemize}
    \item python system\_simulation\_e2.py -c 10 \texttt{-{}-}batch\_size 16 \texttt{-{}-}dataset cifar10 \texttt{-{}-}model resnet18 \texttt{-{}-}pretrained \texttt{-{}-}epochs 100 \texttt{-{}-}rate 0.3

\end{itemize}

\subsubsection{ \textbf{Setting 6: Different Diabetic Retinopathy Datasets}}
\label{appendix_setting6}
Dataset Sources:

\begin{itemize}
    \item Source of Dataset 1, \url{https://www.kaggle.com/competitions/aptos2019-blindness-detection/data}
    \item Source of Dataset 2, \url{https://www.kaggle.com/datasets/mariaherrerot/eyepacspreprocess}
\end{itemize}
To preprocess the dataset download and store the unzipped files in  data/eye\_dataset1 folder and data/eye\_dataset2 folder.
For this create directories using the command:
\begin{itemize}
    \item mkdir data/eye\_dataset1
    \item mkdir data/eye\_dataset2
    
\end{itemize}
The directory structure of data is as follows:
\begin{itemize}
    \item data/eye\_dataset1/train\_images
    \item data/eye\_dataset1/test\_images
    \item data/eye\_dataset1/test.csv
    \item data/eye\_dataset1/train.csv
    \item data/eye\_dataset2/eyepacs\_preprocess/eyepacs\_preprocess/
    \item data/eye\_dataset2/trainLabels.csv
\end{itemize}

Once verify the path of the unzipped folders in the load\_data function of preprocess\_eye\_dataset\_1.py and preprocess\_eye\_dataset\_2.py files.

For Data preprocessing, run the commands mentioned below for both the datasets
\begin{enumerate}

\item python utils/preprocess\_eye\_dataset\_1.py
\item python utils/preprocess\_eye\_dataset\_2.py
\end{enumerate}

   Then use the following commands
\begin{itemize}
      
  \item python PFSL\_DR.py  \texttt{-{}-}pretrained \texttt{-{}-}model resnet18 -c 10  \texttt{-{}-}batch\_size 64 \texttt{-{}-}test\_batch\_size 512 \texttt{-{}-}epochs 50

  \item python  FL\_DR.py -c 10 \texttt{-{}-}batch\_size 64 \texttt{-{}-}test\_batch\_size 512 \texttt{-{}-}epochs 50

  \item python  SL\_DR.py \texttt{-{}-}batch\_size 64 \texttt{-{}-}test\_batch\_size 512 \texttt{-{}-}epochs 50
  \item python  SFLv1\_DR.py  \texttt{-{}-}batch\_size 64 \texttt{-{}-}test\_batch\_size 512 \texttt{-{}-}epochs 50
   \item python SFLv2\_DR.py  \texttt{-{}-}batch\_size 64 \texttt{-{}-}test\_batch\_size 512 \texttt{-{}-}epochs 50
   \end{itemize}

\subsubsection{ \textbf{Test Run on Local System}}
\label{appendix_setting6}

The PFSL script was run on a laptop on cpu with 16 GB RAM for 5 clients each having 50 training datapoints and 200 test datapoints.

The following command was used :
python PFSL\_Setting124 \texttt{-{}-}dataset cifar10 \texttt{-{}-}model resnet18 \texttt{-{}-}pretrained  \texttt{-{}-}setting setting1 \texttt{-{}-}datapoints 50 -c 5 \texttt{-{}-}batch\_size 16 \texttt{-{}-}test\_batch\_size 16.